\documentclass{article}

 \PassOptionsToPackage{numbers, compress}{natbib}

    \usepackage[preprint]{neurips_2024}

\usepackage[utf8]{inputenc} 
\usepackage[T1]{fontenc}    
\usepackage{hyperref}       
\usepackage{url}            
\usepackage{booktabs}       
\usepackage{amsfonts}       
\usepackage{nicefrac}       
\usepackage{microtype}      
\usepackage{xcolor}         

\usepackage{amsmath}
\usepackage{amssymb}
\usepackage{graphicx}
\newcommand{\grey}{\textcolor{gray}}
\newcommand{\mi}{\text{M}(i)}
\newcommand{\oi}{\text{O}(i)}
\newcommand{\mmi}{\text{MM}(i)}
\newcommand{\moi}{\text{MO}(i)}

\newcommand{\ooi}{\text{OO}(i)}

\title{QComp: A QSAR-Based Data Completion Framework for Drug Discovery}

\author{%
  Bingjia Yang\\
  Department of Chemistry\\
  Princeton University\\
  Princeton, NJ 08540 \\
  \And
  Yunsie Chung \\
  Computational and Structural Chemistry \\
  Merck \& Co., Inc \\
  South San Francisco, CA 94080\\
  \And
  Archer Y. Yang \\
  Department of Mathematics and Statistics \\
  McGill University \\ 
  Montreal, Quebec, Canada \\
  \And
  Bo Yuan \\
  Pharmacokinetics, Dynamics, Metabolism, and Bioanalytical \\
  Merck \& Co., Inc. \\
  Rahway, NJ 07065, USA \\
  \And
  Xiang Yu \\
Pharmacokinetics, Dynamics, Metabolism, and Bioanalytical \\
  Merck \& Co., Inc. \\
  Rahway, NJ 07065, USA \\
}
\begin{document}

\maketitle

\begin{abstract}

  In drug discovery, in vitro and in vivo experiments reveal biochemical activities related to the efficacy and toxicity of compounds. The experimental data accumulate into massive, ever-evolving, and sparse datasets. Quantitative Structure-Activity Relationship (QSAR) models, which predict biochemical activities using only the structural information of compounds, face challenges in integrating the evolving experimental data as studies progress.
  We develop QSAR-Complete\footnote{The code is available at https://github.com/iceplussss/QSAR-Complete} (QComp), a data completion framework to address this issue. Based on pre-existing QSAR models, QComp utilizes the correlation inherent in experimental data to enhance prediction accuracy across various tasks. Moreover, QComp emerges as a promising tool for guiding the optimal sequence of experiments by quantifying the reduction in statistical uncertainty for specific endpoints, thereby aiding in rational decision-making throughout the drug discovery process.
\end{abstract}

\section{Introduction}

Quantitative Structure-Activity Relationship (QSAR) modeling is one of the most important approaches for data-driven prediction of molecular properties~\cite{von1998encyclopedia, svetnik2003random, noble2006support, obrezanova2007gaussian}, with recent progress led by deep learning~\cite{ma2015deep, gawehn2016deep, gilmer2017neural, Yang2019AnalyzingPrediction, huang2020deeppurpose, tropsha2023integrating}. Sophisticated deep learning methods can model various chemical properties with a unified (multi-task) neural network model~\cite{ma2015deep, dahl2014multi, kearnes2016modeling, Wenzel2019PredictiveSets, feinberg2020improvement}. 

QSAR finds major applications in material and drug discovery~\cite{muratov2020qsar, tropsha2023integrating}. It is the de facto method for in silico high-throughput screening ~\cite{jia2022machine, feinberg2020improvement} of a database of molecules with unknown properties. Its dominance is partially due to its simplicity: only the structure of a molecule is required for predicting molecular properties. This simplicity, however, becomes less desirable in stages past in silico modeling, where QSAR models face challenges in effectively incorporating newly acquired measurements towards improved prediction~\cite{walter2022analysis}. 
One potential solution involves retraining the multi-task QSAR model with both the original training set and the newly acquired data. The effectiveness of such retraining is, however, questionable when the newly acquired data is negligible compared to the size of the original training set, a common scenario in industrial practice of material and drug discovery due to the high cost of new experiments and the massive size of historical data. Retraining a large deep learning model for minor data updates is also uneconomical. Therefore, a data completion method that can effectively leverage pre-existing QSAR models at a low cost is desirable. 

For this purpose, we develop a QSAR-based data completion framework, named ``QSAR-Complete'' or ``QComp'' for brevity. QComp treats chemical activities $\mathbf{y}$ of a molecule as a probability distribution $\mathcal{P}(\mathbf{y} | \mathbf{x} )$ decided by the chemical descriptor $\mathbf{x}$ of the molecule. 
Typical structure-based QSAR models can be understood as to directly predict $\text{argmax}_\mathbf{y} \mathcal{P}(\mathbf{y} | \mathbf{x} )$ as a function of $\mathbf{x}$. QComp addresses instead the case that some entries of $\mathbf{y}$ are determined already by experimental data. To do so, QComp parameterizes the probability distribution of the missing entries of $\mathbf{y}$ as a function of known entries and $\mathbf{x}$. The maximum of such a function yields optimal data completion. Moreover, QComp incorporates a pre-existing QSAR model in a natural way, such that QComp can reproduce the structure-based  QSAR prediction when $\mathbf{y}$ is entirely unknown. 
We demonstrate the application of QComp in modeling absorption, distribution, metabolism, elimination, and toxicity (ADMET) for small molecules and peptides because these properties are tightly bound to the efficacy and safety of drug candidates. We also apply QComp to the optimization of decision-making in drug discovery. More applications are expected in other material and drug discovery tasks facing similar challenges.

We summarize our main contributions below: 
\begin{itemize}
    \item We propose the QComp approach that leverages any QSAR model for more accurate data completion, by exploiting the correlation between endpoints. 
    \item We demonstrate that QComp systematically improves upon structure-based QSAR for ADMET data completion. Moreover, 
    QComp shows advantages in accuracy, robustness and interpretability, compared to several standard data completion methods fed with the same side information from the existing QSAR model. 
    \item We show that QComp can guide the rational design of the sequence of in vivo and in vitro experiments carried out in drug discovery, by optimizing the marginal utility. 
\end{itemize}


\section{Related works}

Over past decades, numerous general imputation algorithms~\cite{troyanskaya2001missing, oba2003bayesian, liew2011missing, kim2005missing, van1999flexible, stekhoven2012missforest, simm2015macau} have been proposed. 
For example, multivariate Imputation by Chained Equations (MICE) ~\cite{van1999flexible} and MissForest~\cite{stekhoven2012missforest} are leading members in the category of iterative imputers. They model each feature as a function of others, starting by replacing missing values with statistical means or the most frequent values. Then, the imputed entries are updated iteratively in a round-robin fashion. 
Another major category is matrix factorization-based methods~\cite{koren2009matrix}. Macau~\cite{simm2015macau}, a member of this category, has been applied to ADMET tasks~\cite{walter2022analysis}.
Unlike QComp, these general algorithms do not base data completion on another predictive model. However, they are flexible enough to incorporate additional information for improved performance on sparse datasets, which allows fair comparison with QComp.

In addition to general methods, specific data completion methods have been tailored for predicting chemical properties, such as Alchemite~\cite{whitehead2019imputation} and pQSAR~\cite{Martin2017}. Alchemite, as an iterative imputer, updates imputed values through a multi-task neural network with chemical descriptors and activities as input. Here, directly utilizing a neural network for imputation raises concerns about convergence~\cite{oberman2020missing, nguyen2021practical}, a typical issue for iterative imputers. 
The risk of divergence is certain for a deep neural network that often experiences overfitting and unreliable extrapolation on insufficiently large datasets - a common scenario for in vivo ADMET properties. The pQSAR model, also as an iterative imputer, avoids a divergent imputation by using very few iterations (up to nine in Ref.~\cite{Martin2017, kim2020extension}), which, however, potentially leads to sub-optimal imputation.
Due to these challenges in iterative imputation, our QComp approach instead builds data completion on a probabilistic framework with a well-defined optimum. 

\section{Methods}\label{methods}

\subsection{Probabilistic framework of QComp}
For a molecule uniquely labeled as $i$ in a molecular database $\mathcal{I}$, let $\mathbf{x}^{(i)}$ be its chemical descriptor and the row vector $\mathbf{y}^{(i)}=(y_1^{(i)}, y_2^{(i)}, \cdots, y_p^{(i)})$ represents its $p$ (chemical) activities/assays. 
We use $\mathbf{y}^{\oi}$ to denote the sub-vector (of length $p^{(i)}_\text{O}$) of $\mathbf{y}^{(i)}$ containing those known (observed) activities from experiments, and $\mathbf{y}^{\mi}$ the sub-vector (of length $p^{(i)}_\text{M}=p-p^{(i)}_\text{O}$) containing unknown (missing) activities as stochastic variables. The partition $\mathbf{y}^{(i)}=(\mathbf{y}^{\mi}, \mathbf{y}^{\oi})$ varies for different $i\in \mathcal{I}$. 

The task of QComp is to determine $\mathcal{P}(\mathbf{y}^{\mi}|\mathbf{y}^{\oi},\mathbf{x}^{(i)})$ as a conditional distribution of $\mathcal{P}(\mathbf{y}^{(i)}|\mathbf{x}^{(i)})$. The optimal data completion is the conditional expectation $\tilde{\mathbf{y}}^{\mi} = \mathbb{E} (\mathbf{y}^{\mi} | \mathbf{y}^{\oi}, \mathbf{x}^{(i)})$. Meanwhile, the conventional QSAR model gives access to an estimation of $\text{argmax}_\mathbf{y} \mathcal{P}(\mathbf{y}^{(i)} | \mathbf{x}^{(i)})$ as a function of $\mathbf{x}^{(i)}$, denoted by  
 $\boldsymbol{f}^{(i)}=(f_1(\mathbf{x}^{(i)}), f_2(\mathbf{x}^{(i)}), \cdots, f_p(\mathbf{x}^{(i)}))$. QComp utilizes this estimation and assumes that $\mathbf{y}^{(i)}$ conditional on $\mathbf{x}^{(i)}$ follows a multivariate Gaussian distribution $\mathbf{y}^{(i)}|\mathbf{x}^{(i)}\sim N(\boldsymbol{\mu}^{(i)}, \boldsymbol{\Sigma})$ with the probability density function
$\mathcal{P}(\mathbf{y}^{(i)}| \mathbf{x}^{(i)})$ 
given by 
\begin{equation}\label{dist}
    \mathcal{P}(\mathbf{y}^{(i)} | \mathbf{x}^{(i)} )  = \big( (2\pi)^p |\boldsymbol{\Sigma}|\big)^{-\frac{1}{2}}  \text{exp} \Big( -\frac{1}{2} \left( \mathbf{y}^{(i)} - \boldsymbol{\mu}^{(i)} \right) \boldsymbol{\Sigma}^{-1} \left( \mathbf{y}^{(i)} - \boldsymbol{\mu}^{(i)}\right)^{\top} \Big).
\end{equation}
This is not to be confused with assuming the activity $\mathbf{y}^{(i)}$ itself is normally distributed (see Sec. \ref{validation_of_assumption} for details). 
The row vector $\boldsymbol{\mu}^{(i)} = \boldsymbol{f}^{(i)}\mathbf{B} + \mathbf{b}$ is a linear transformation of the QSAR prediction $\boldsymbol{f}^{(i)}$, serving as a multi-task calibration of given QSAR models. $\mathbf{B}$ is a $p\times p$ matrix, and $\mathbf{b}$ a $1 \times p$ vector. The covariance matrix $\boldsymbol{\Sigma}$ is a positive-definite $p\times p$ matrix. $|\boldsymbol{\Sigma}|$ denotes its determinant.  Specifically, $\boldsymbol{\Sigma}$ is represented by its Cholesky decomposition and only the resulting lower triangle matrix is treated as free parameters.  In the following, we use $\theta$ to represent the group of parameters determining $\mathbf{B}$, $\mathbf{b}$ and $\boldsymbol{\Sigma}$. 

For each $i$ and the partition $\mathbf{y}^{(i)}=(\mathbf{y}^{\mi}, \mathbf{y}^{\oi})$. The calibrated QSAR prediction $\boldsymbol{\mu}^{(i)}$ can be correspondingly partitioned as $(\boldsymbol{\mu}^{\mi}, \boldsymbol{\mu}^{\oi})$. And $\boldsymbol{\Sigma}$ can be partitioned as the block matrix  $\Big(\begin{smallmatrix}
        \boldsymbol{\Sigma}^{\mmi} & \boldsymbol{\Sigma}^{\moi}  \\
        [\boldsymbol{\Sigma}^{\moi}]^\top & \boldsymbol{\Sigma}^{\ooi} 
\end{smallmatrix}\Big)$.
Here, $\boldsymbol{\Sigma}^{\mmi}$ represents the $p_\text{M}^{(i)}\times p_\text{M}^{(i)}$ submatrix of $\boldsymbol{\Sigma}$ associated with the covariance of $\mathbf{y}^{\mi}$. The meaning of $\boldsymbol{\Sigma}^{\moi}$ and $\boldsymbol{\Sigma}^{\ooi}$ should be self-evident. 

 
\subsection{Training}
Within QComp, the likelihood of the observation $\mathbf{y}^{\oi}$ follows the marginal Gaussian distribution
\begin{equation}
\small
\mathcal{P}(\mathbf{y}^{\oi} |\mathbf{x}^{(i)}) = \int \mathcal{P}(\mathbf{y}^{(i)}| \mathbf{x}^{(i)}) d\mathbf{y}^{\mi} = \frac{\text{exp}\Big( 
-\frac{1}{2} \left( \mathbf{y}^{\oi} - \boldsymbol{\mu}^{\oi} \right) (\boldsymbol{\Sigma}^{\ooi})^{-1} \left( \mathbf{y}^{\oi} - \boldsymbol{\mu}^{\oi}\right)^{\top} 
\Big)}
{\sqrt{(2\pi)^{p^{(i)}_\text{O}} |\boldsymbol{\Sigma}^{\ooi}|}}.
\end{equation} 
We define the following log-likelihood loss function with respect to $\theta=(\mathbf{B}, \mathbf{b}, \boldsymbol{\Sigma})$:
\begin{equation}
\ell(\theta) = -\log \prod_{i\in \mathcal{I}} \mathcal{P}(\mathbf{y}^{\oi} |\mathbf{x}^{(i)}) = -\sum_{i\in\mathcal{I}} \log \mathcal{P}(\mathbf{y}^{\oi} |\mathbf{x}^{(i)}).     
\end{equation}
$\widehat{\theta}=(\widehat{\mathbf{B}}, \widehat{\mathbf{b}}, \widehat{\boldsymbol{\Sigma}})$ denotes the optimal values of $\theta$, defined as 
$\widehat{\theta} = \underset{\theta}{\arg\min}
 \ \ell(\theta)$.
This optimization problem can be solved by carrying out gradient descent on $\theta$.  

\subsection{Data completion}
After fixing $\widehat{\theta}$, a QComp model can be used for one-shot data completion. Note that $\mathbf{y}^{\mi}$ conditioned on $\mathbf{y}^{\oi}$ follows a Gaussian distribution, i.e. 
\begin{equation}
\mathbf{y}^{\mi} | \mathbf{y}^{\oi}, \mathbf{x}^{(i)}\sim N(\tilde{\boldsymbol{\mu}}^{\mi}, \widetilde{\boldsymbol{\Sigma}}^{\mmi}),   
\end{equation}
 where 
 \begin{equation}
(\tilde{\boldsymbol{\mu}}^{\mi})^{\top} = (\boldsymbol{\mu}^{\mi})^{\top} + \widehat{\boldsymbol{\Sigma}}^{\moi} (\widehat{\boldsymbol{\Sigma}}^{\ooi})^{-1} (\mathbf{y}^{\oi} - \boldsymbol{\mu}^{\oi})^{\top}
\end{equation}
and 
\begin{equation}
\widetilde{\boldsymbol{\Sigma}}^{\mmi} = \widehat{\boldsymbol{\Sigma}}^{\mmi} - \widehat{\boldsymbol{\Sigma}}^{\moi} (\widehat{\boldsymbol{\Sigma}}^{\ooi})^{-1} [\widehat{\boldsymbol{\Sigma}}^{\moi}]^\top.   
\end{equation}
The corresponding probability density function is
\begin{equation}
      \mathcal{P}(\mathbf{y}^{\mi} | \mathbf{y}^{\oi}, \mathbf{x}^{(i)}) = \frac{\text{exp}\Big( 
-\frac{1}{2} \left( \mathbf{y}^{\mi} - \tilde{\boldsymbol{\mu}}^{\mi} \right) (\widetilde{\boldsymbol{\Sigma}}^{\mmi})^{-1} \left( \mathbf{y}^{\mi} - \tilde{\boldsymbol{\mu}}^{\mi}\right)^{\top} 
\Big)}
{\sqrt{(2\pi)^{p^{(i)}_\text{M}} |\widetilde{\boldsymbol{\Sigma}}^{\mmi}|}},
\end{equation}
The optimal data completion given by QComp for the missing assays is therefore 
\begin{equation}
\mathbb{E}(\mathbf{y}^{\mi} | \mathbf{y}^{\oi}, \mathbf{x}^{(i)})=\tilde{\boldsymbol{\mu}}^{\mi}
\end{equation}




A comment on the data completion uncertainty is in order. Here,
the uncertainty related to $\tilde{\boldsymbol{\mu}}^{\mi}$ is not simply the diagonal of $\widetilde{\boldsymbol{\Sigma}}^{\mmi}$, unless one can ignore the uncertainty embedded in the QSAR prediction, which is usually far from negligible. We construct a composite uncertainty in the Appendix.~\ref{sec_composite_uncertainty} to address this extra complication. However, even without further construction, here we are already able to have a clear idea of how much certainty one can gain on missing assays $\mathbf{y}^{\mi}$ by knowing the experimental measurements $\mathbf{y}^{\oi}$. The gain of certainty is simply  the diagonal terms in $ \boldsymbol{\Sigma}^{\moi} (\boldsymbol{\Sigma}^{\ooi})^{-1} [\boldsymbol{\Sigma}^{\moi}]^\top$.

\section{Experiments}\label{experiments}

\subsection{Data and model details}\label{sec_data}

\paragraph{Datasets} We apply our approach to three proprietary ADMET datasets and one public ADMET dataset. 
The first proprietary dataset (ADMET-750k dataset) contains sparse data of 32 in vitro and in vivo ADMET assays for around 750000 small molecules. 
The second proprietary dataset (fup dataset) is a three-assay sparse dataset for fraction unbound in plasma data. 
The third proprietary dataset (peptide dataset) contains sparse data of 26 ADMET assays for peptides. 
The public dataset contains data of 25 ADMET assays for 114112 small molecules. The details of these datasets, including the list of chemical activities and the Pearson correlation between activities, can be found in Appendix \ref{sec_data_details}.
We will benchmark QComp on the largest ADMET-750k dataset, which is accumulated from consistent industrial drug discovery practices. Similar benchmarking procedure is performed for the small public dataset, which is compiled from various public sources\cite{Wenzel2019PredictiveSets, Iwata2022PredictingData, Kim2023PubChemUpdate, Watanabe2018PredictingRanges, Falcon-Cano2022ReliableApproaches, Esposito2020CombiningSubstrates, Braga2015Pred-hERG:Toxicity, Aliagas2022ComparisonSets, Perryman2020PrunedSolubility, Meng2022BoostingCuration, Vermeire2022PredictingTemperatures}, for reproducibility of the QComp approach (see Appendix \ref{sec_public}).

\paragraph{Base QSAR models} For ADMET-750k and the public dataset, we train multi-task Chemprop models as the base QSAR (see Appendix for details). Chemprop model utilizes a directed message-passing neural network (D-MPNN) to predict molecular properties based on the graph representation of molecules~\cite{Yang2019AnalyzingPrediction, Heid2024}. For the fup and peptide datasets, random forest models are used as base QSAR models~\cite{svetnik2003random} deployed at Merck.

\paragraph{Baseline data completion models} We compare the QComp approach with three baseline data completion methods: MICE~\cite{van1999flexible}, Macau~\cite{simm2015macau}, and Missforest~\cite{stekhoven2012missforest}.
We provide the three baseline methods with the same QSAR predictions accessed by QComp. Specifically, for MICE and MissForest, we extend the dataset by appending QSAR predictions as supplementary columns. For example, the ADMET-750k dataset, originally containing 32 assay columns, is extended to 64 columns, where the extra 32 columns are Chemprop predictions with no missing value.
For Macau, we use the QSAR predictions as side information~\cite{simm2015macau, de2018effect}. The parameters for these methods are provided in the Appendix.

\paragraph{Data splitting strategies} For the ADMET-750k dataset, during the training of the Chemprop models and the QComp model, we split the entire dataset into 90\% training/validation and 10\% test subsets using an assay-based temporal split, such that the test set contains the most recent assay data. 
Compound-based temporal splitting ~\cite{walter2022analysis} of the same dataset is also carried out for comparison, leading to a similar performance of QComp (see Appendix ~\ref{sec_dataset1_details}). 
For the other three datasets, we perform only 10\% random splitting as the assay measurement date information is not available.

\subsection{Validation of assumption}\label{validation_of_assumption}

\begin{figure}[h!]
    \centering
    \includegraphics[width=0.9\linewidth]{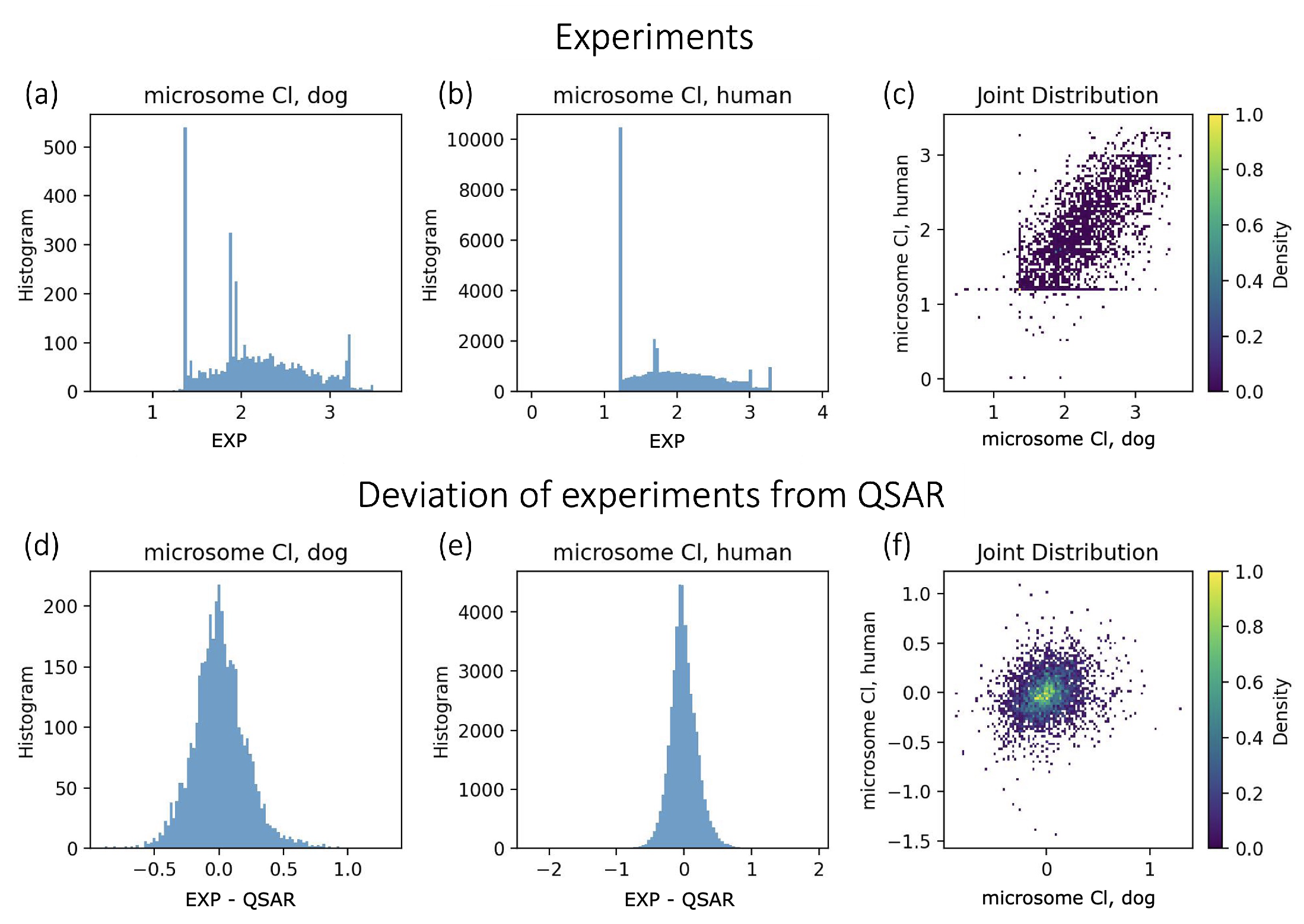}
    \caption{
    (a,b) Histograms of the ``microsome Cl'' assays for dogs and humans. (c) The heatmap of the joint distribution of ``microsome Cl, dog'' and ``microsome Cl, human''. (d,e) Histograms of the deviation of ``microsome Cl'' assays from the QSAR predictions. (f) The heatmap of the joint distribution associated with the quantities in (d) and (e).  
}\label{fig_distribution_microsomeCl}
\end{figure}

Here, we examine the basic assumption of QComp --- the deviation of the experimental value of an assay from the QSAR prediction is distributed normally (see Eq.~\ref{dist}). Evidently, the assumed distribution is subject to the quality of the QSAR model. For a trivial QSAR model that gives constant predictions independent of chemical descriptors, the distribution of $(\mathbf{y}^{(i)} - \boldsymbol{\mu}^{(i)})$ can be far from being Gaussian. This is exemplified by Fig.~\ref{fig_distribution_microsomeCl}(a,b), where we show with histograms the plain distribution of the experimental values of two assays, ``microsome Cl dog'' and ``microsome Cl human'', in ADMET-750k dataset. For both assays, the peak of the histogram is located near the lower end of the distribution, in sharp contrast to a typical Gaussian distribution. Furthermore, the joint distribution of the two assays (Fig.~\ref{fig_distribution_microsomeCl}(c)) is not close to a 2D Gaussian distribution. 

The situation is different when the QSAR model is properly trained. We examine the multi-task Chemprop model (trained on the same dataset) that serves as the base $\boldsymbol{\mu}^{(i)}$. Fig.~\ref{fig_distribution_microsomeCl}(d) (Fig.~\ref{fig_distribution_microsomeCl}(e)) shows with histogram the distribution of the ``microsome Cl, dog (human)'' component of $(\mathbf{y}^{(i)} - \boldsymbol{\mu}^{(i)})$. The distributions display a close resemblance to the 1D Gaussian distribution centered at zero. Meanwhile, Fig.~\ref{fig_distribution_microsomeCl}(f) shows that the joint distribution of the two assays is similar to a zero-centered 2D Gaussian distribution with positive off-diagonal covariance. The non-zero off-diagonal covariance, i.e. the correlation between different assays, is what to be utilized by QComp to exceed the capability of bare QSAR. Of course, not all pairs of assays display non-zero off-diagonal covariance, since two chemical properties can not always be statistically correlated. 

Besides the two assays used as examples here, other pairs of assays in all our datasets yield similar results. These observations validate the assumption of QComp over our datasets. However, we acknowledge the possibility that the assumption may fail in cases where the QSAR models are suboptimal.

\subsection{Benchmarking QComp for ADMET data completion: ADMET-750k dataset}\label{sec_dataset1}

We benchmark QComp on the ADMET-750k dataset with the multi-task Chemprop model as the base QSAR model. QComp, MICE, Missforest, and Macau models are trained on the same training set with the QSAR predictions from Chemprop as side information. 
Then, these methods are evaluated on the test set with the following protocol.  For any assay-$i$, we mask the column of assay-$i$ in the test set as totally missing and complete this column with all other columns. The completed column is then compared against available experimental data of assay-$i$ with the squared Pearson correlation coefficient $r^2$ as a metric. 




\begin{table}[h!]\small
    \centering
    \caption{$r^2$ scores of QComp, Missforest, Macau, MICE, and the base QSAR model on ADMET-750k dataset with assay-based temporal splitting. For each assay, the highest $r^2$ score is marked in \textbf{bold}. The second highest $r^2$ score is marked in \grey{\textbf{bold and grey}}.}
    \label{tab_dataset1}
\begin{tabular}{l lllll}
\toprule
Assay name & QComp & Missforest & Macau & MICE & Chemoprop \\
\midrule
Papp & \textbf{0.751} & 0.736 & 0.731 & \textbf{0.751} & \grey{\textbf{0.749}} \\
CaV 1.2 & \textbf{0.361} & 0.333 & 0.346 & \grey{\textbf{0.359}} & 0.346 \\
NaV 1.5 & \textbf{0.364} & 0.338 & 0.315 & \textbf{0.364} & \grey{\textbf{0.358}} \\
Cl, dog & \textbf{0.509} & 0.273 & 0.220 & \grey{\textbf{0.422}} & 0.276 \\
Cl, rat & \textbf{0.992} & 0.836 & 0.550 & \grey{\textbf{0.967}} & 0.560 \\
hepatocyte Cl, dog & \textbf{0.664} & 0.520 & 0.496 & \grey{\textbf{0.626}} & 0.534 \\
microsome Cl, dog & \grey{\textbf{0.543}} & 0.442 & 0.451 & \textbf{0.565} & 0.441 \\
hepatocyte Cl, human & \textbf{0.570} & 0.467 & 0.460 & \grey{\textbf{0.540}} & 0.482 \\
microsome Cl, human & \textbf{0.695} & 0.600 & 0.577 & \grey{\textbf{0.657}} & 0.595 \\
hepatocyte Cl, rat & \textbf{0.554} & 0.413 & 0.441 & \grey{\textbf{0.537}} & 0.421 \\
microsome Cl, rat & \textbf{0.724} & 0.608 & 0.611 & \grey{\textbf{0.705}} & 0.613 \\
CYP2C8 & \textbf{0.469} & 0.421 & 0.423 & \grey{\textbf{0.467}} & 0.457 \\
CYP2C9 & \textbf{0.341} & 0.302 & 0.317 & \textbf{0.341} & \grey{\textbf{0.328}} \\
CYP2D6 & \textbf{0.316} & 0.119 & 0.286 & \textbf{0.316} & \grey{\textbf{0.297}} \\
CYP3A4 & \textbf{0.466} & 0.440 & 0.443 & \grey{\textbf{0.461}} & 0.451 \\
CYP,TDI,3A4,ratio & \textbf{0.134} & 0.116 & 0.009 & \grey{\textbf{0.133}} & 0.132 \\
EPSA & \grey{\textbf{0.834}} & 0.813 & 0.511 & 0.815 & \textbf{0.836} \\
halflife, dog & \textbf{0.784} & \grey{\textbf{0.750}} & 0.401 & 0.543 & 0.413 \\
halflife, rat & \textbf{0.772} & \grey{\textbf{0.721}} & 0.338 & 0.522 & 0.245 \\
hERG MK499 & \textbf{0.500} & 0.475 & 0.495 & 0.497 & \grey{\textbf{0.499}} \\
Fu,p, human & \grey{\textbf{0.698}} & 0.668 & \textbf{0.708} & 0.669 & 0.693 \\
LogD & \textbf{0.901} & 0.886 & \grey{\textbf{0.900}} & \grey{\textbf{0.900}} & \grey{\textbf{0.900}} \\
PAMPA & \textbf{0.743} & 0.492 & 0.523 & 0.011 & \grey{\textbf{0.732}} \\
PXR activation & 0.433 & 0.419 & 0.432 & \grey{\textbf{0.434}} & \textbf{0.435} \\
Fu,p, rat & \textbf{0.717} & 0.654 & 0.637 & \grey{\textbf{0.696}} & 0.671 \\
Fassif Solub & \grey{\textbf{0.493}} & 0.454 & 0.383 & \textbf{0.498} & 0.415 \\
Vd, rat & \textbf{0.993} & 0.815 & 0.633 & \grey{\textbf{0.959}} & 0.622 \\
MRT, dog & \textbf{0.926} & 0.858 & 0.488 & \grey{\textbf{0.920}} & 0.433 \\
MRT, rat & \textbf{0.995} & 0.695 & 0.264 & \grey{\textbf{0.992}} & 0.233 \\
SOLY7 & \textbf{0.703} & \grey{\textbf{0.680}} & 0.637 & 0.662 & 0.647 \\
PGP, rat & \textbf{0.590} & 0.565 & 0.576 & \grey{\textbf{0.589}} & 0.585 \\
PGP, human & \grey{\textbf{0.435}} & 0.122 & 0.000 & 0.025 & \textbf{0.446} \\
\bottomrule
\end{tabular}
\end{table}

The $r^2$ score obtained by the four data completion methods on the test set is reported in Table.~\ref{tab_dataset1}. 
Overall, the base QSAR model achieves a mean $r^2$ score (averaged over all 32 assays) of 0.487. QComp, MICE, Missforest, and Macau achieve a mean $r^2$ score of 0.620, 0.555, 0.526, and 0.447, respectively. 
QComp outperforms other methods by a large margin, with a 27\% improvement over the base. 
Although not reported in Table.~\ref{tab_dataset1}, we have calculated the standard deviation of the $r^2$ scores obtained by QComp as an error bar, resulting from random initialization of $\boldsymbol{\Sigma}$. All error bars are of the order of 0.001, which is negligible compared to the improvement achieved by QComp on the mean $r^2$ score. 
Then, to examine the $r^2$ score on the individual assay, we consider a simple criterion:  a successful data completion method should not reduce the $r^2$ score from the base QSAR model by more than 0.01. 
QComp meets the requirement for all assays except ``PGP, human'', where QComp deviates from the base QSAR model by merely 0.011. 
In contrast, all other methods can yield $r^2$ scores significantly lower than the base. ``PGP, human'' and  ``PAMPA'' are outstanding examples where other data-completion methods reduce the base $r^2$ score in the order of 0.1. The comparison shows the excellent robustness of QComp.
Moreover, QComp outperforms other methods for all assays but ``microsome Cl, dog'', ``Fu,p, human'', ``Fassif Solub'', where QComp loses by a small margin.  Nevertheless, for some assays, such as ``Papp'' and ``NaV 1.5'', the improvement brought by QComp has no statistical significance. 
Meanwhile, Macau typically underperforms all other methods including the base QSAR model. A possible explanation is that Macau assumes a low-dimensional representation of the data matrix, which is not justified for the ADMET dataset.  
Comparing QComp, MICE, and Missforest, the success of QComp may be due to its constrained way of utilizing the correlation among ADMET properties: the simple Gaussian model adopted by QComp disregards non-linear correlations and greatly reduces over-fitting. 
This drastic simplification, however, should not impair much the capability of QComp since we are modeling the deviation of assay from QSAR predictions. The non-linear correlation between assays has been captured by the non-linear base QSAR model. 
The importance of the base QSAR model can also be seen from another perspective: the mean  $r^2$ score obtained by MICE, Missforest, and Macau will be significantly smaller if we do not provide QSAR predictions as side information.


Last, note that the simple benchmarking protocol adopted disregards the complication that the location of missing entries is not randomly distributed.  In practice, correlated assays from the same experiment will be simultaneously present or missing. Here, ``MRT'', ``halflife'', ``Cl'', and ``Vd'' assays of the same animal come out of the same experiment.  QComp yields unrealistically large improvement upon the base QSAR model for these assays. Although demonstrating how effective QComp is to utilize assay-assay correlation, such improvement should not be expected in practice. In Appendix ~\ref{sec_masked_cols}, we test QComp in a more realistic setting. For completing any assay-$i$, we mask the columns of assays from the same experiment as assay-$i$'s. Then the improvement of QComp upon base QSAR model becomes reasonable for ``MRT'', ``halflife'', ``Cl'', and ``Vd''.

\subsection{Enhancing prediction of human assay with animal data: fup dataset}\label{sec_dataset2}

Here we demonstrate how QComp improves predictions of Human assays based on data obtained from animal experiments. This is a major incentive for deploying data completion frameworks in drug discovery.

The fup dataset contains three assays crucial for indicating drug efficacy: fraction unbound in plasma (fup) of rat, dog, and human. 
We train the QComp model on the training set with single-task random forest models as base QSAR models.
To illustrate exactly how much the prediction of the human assay benefits from animal data, we extract from the test set a subset, where the rat, dog, and human fup data are all present. 
For this subset, the Pearson $r^2$ score obtained by the base QSAR model on human fup is only 0.494. For comparison, if we mask both dog and human fup in the subset, but keep rat data visible, the Pearson $r^2$ score obtained by QComp on completing human fup is 0.729. Similarly, masking dog data instead of rat data, the Pearson $r^2$ score obtained by QComp on human fup is 0.742. If we keep all dog and rat data unmasked, QComp obtains a Pearson $r^2$ score of 0.751 for predicting human fup. The error bar of these $ r^2$ scores, resulting from random initialization of $\boldsymbol{\Sigma}$, is again of the order of 0.001.
These results suggest that knowledge of either rat or dog data can significantly improve the prediction of human fup, displayed as a nearly $50\%$ increment in $r^2$ score. Interestingly, knowing both rat and dog fup brings no substantial new information for human fup compared to knowing only rat or dog fup.

 
Here, QComp shows the capability of exploiting the correlation between human and animal fup, in contrast to conventional QSAR models. We suggest that an efficient way of predicting human fup, while experiments on humans are not available, is to measure either dog or rat fup and adopt the QComp framework for data completion.

\subsection{Data completion beyond small-molecule activities: peptide dataset}\label{sec_dataset3}
It is believed that peptides hold immense therapeutic potential. Hence, accurate prediction of peptide properties is no less important than predicting properties of small molecules. 
Here, we focus on the peptide dataset containing 26 chemical properties of peptides (see Sec.~\ref{sec_peptice_dataset}). We establish QComp on random forest QSAR models previously trained on the same dataset. 
\begin{figure}[bth]
    \centering
    \includegraphics[width=\textwidth]{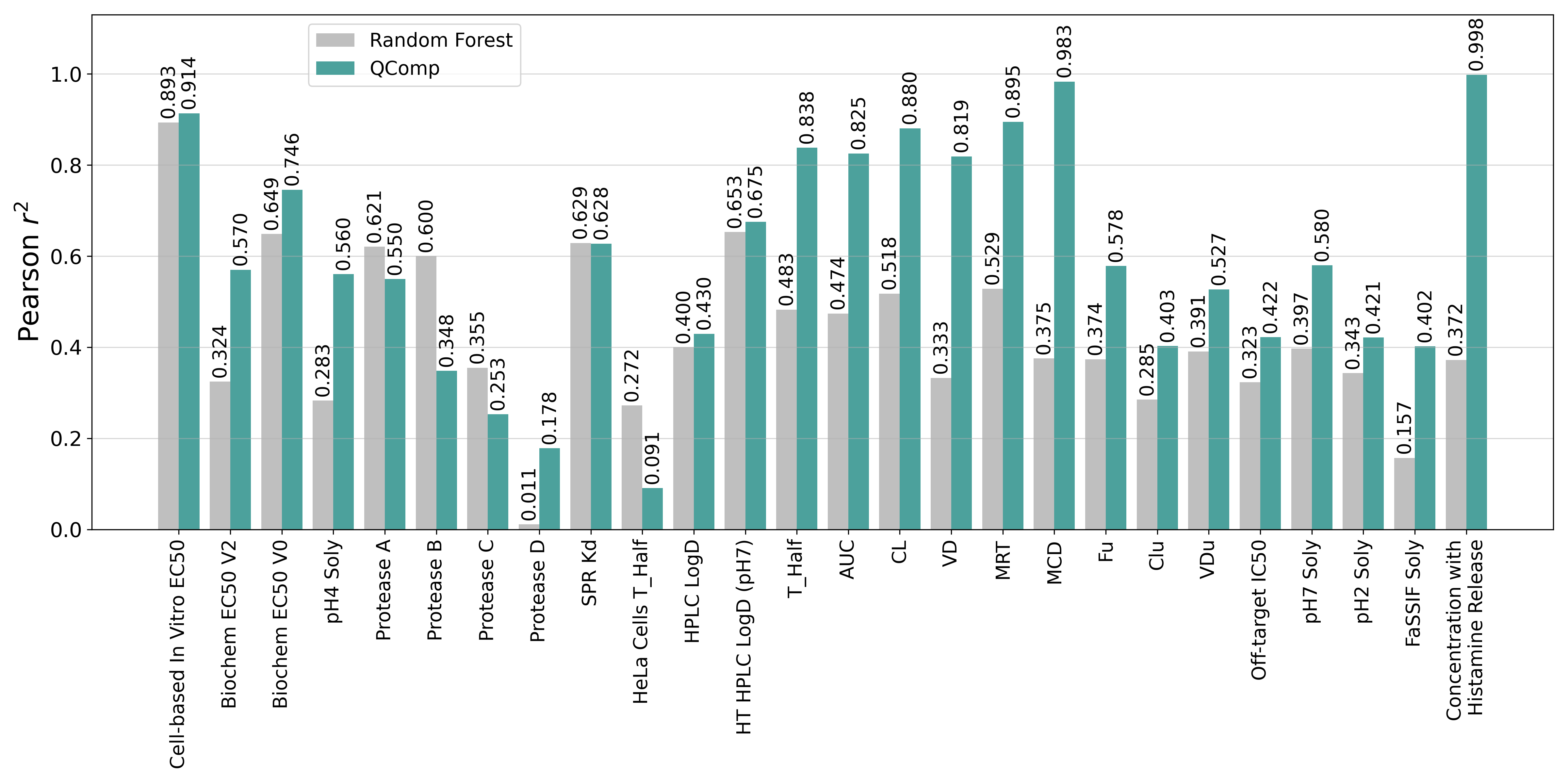}
    \caption{ $r^2$ scores of QComp and the base (random forest) QSAR model on the peptide dataset with random splitting.
    }
    \label{fig_peptide_r2}
\end{figure}

To evaluate the performance of QComp, we engage the same benchmarking protocol used in Sec.~\ref{sec_dataset1}. The $r^2$ scores obtained by the base QSAR model and QComp are plotted in Fig.~\ref{fig_peptide_r2}. The error bars resulting from random initialization are still of the order of 0.001, hence not reported. 
QComp improves upon or maintain the accuracy of base QSAR model on 22 assays. The remaining 4 assays are ``Protease A/B/C'' and ``Hela Cells T\_half''.  The cause of the anomaly is identified as the insufficient number of data. Specifically, ``Protease B/C/D'' and ``Hela Cells T\_half'' are the four assays with the smallest number of experimental data in the dataset (The exact number of data is not reported since they are proprietary information).
The anomalies include not only underperforming base QSAR, as in the case of ``Protease B/C'' and ``Hela Cells T\_half'', but also outperforming base QSAR by an unrealistic margin, as in the case of ``Protease D'' where QComp increases $r^2$ by more than ten times. Simultaneous, the anomaly of ``Protease A'' can be accounted for. As already suggested by its name, ``Protease A'' is highly correlated with ``Protease B/C/D'' (see Fig.~\ref{fig_peptide_corr}). The completion of the former is thus highly sensitive to the deviation of the experimental data of the latter from QSAR values, which leads to unreasonable predictions of ``Protease A'' for some peptides. Therefore, the base QSAR model outperforms QComp by a small margin on ``Protease A''.

Now the anomalies have been accounted for by the lack of data, which in principle should be avoided by all statistical learning algorithms, we turn to the 22 assays where QComp is favorable. The average $r^2$ score of these assays, excluding ``Protease D'', is raised from 0.428 to 0.673 by QComp. 
A more detailed statistical analysis of the results can be carried out in a way similar to what has been reported by Sec.~\ref{sec_dataset1}, hence omitted here. 

Here, we extend the application of QComp beyond the scope of small molecules. The assumption of normally distributed data deviation from base QSAR predictions still holds for the peptide dataset. QComp yields systematic improvement upon the base random forest models. 
 
\subsection{Rational decision-making with QComp}

When QComp predicts a missing assay, it also gives the gain of certainty (GOC) brought by the available experimental data. GOC quantifies the reduction in statistical uncertainty of a QComp prediction compared to the corresponding base QSAR prediction. In practice, GOC can be used as an indicator of how effective a data completion is. 

Within our framework, GOC is a statistical quantity that does not depend on the chemical descriptors of individual compounds. Specifically, for imputing a missing assay-$k$ of an arbitrary compound, the GOC depends only on the indices of the other assays with available experimental data for this compound.  
This allows a convenient greedy scheme for the decision-making procedure in experimental ADMET studies. 

We consider the scenario that the assay-$k$ is of primary interest for a new compound with no experimental data yet. We assume the direct measurement of assay-$k$ is expensive. For example, assay-$k$ is an in vivo property. The goal here is to measure a few in vitro assays instead and impute in vivo assay-$k$ with the acquired in vitro data and the pre-existing QSAR prediction. For such circumstances, we propose a scheme that predicts the sequence of in vitro assays to be measured for maximizing short-term gain.  
The scheme first prioritizes the measurement of the in vitro assay-$k_0$ that brings the highest GOC for assay-$k$. Then, after assay-$k_0$ gains experimental data, the GOC for assay-$k$ with respect to the measurement of other in vitro assays changes. One can re-calculate the GOC and prioritize again the assay that brings the highest GOC for assay-$k$. This procedure repeats until the GOC for assay-$k$ is ignorable for any remaining missing in vitro assay, meaning we can not significantly improve the quality of data completion anymore.

\begin{figure}[h!]
    \centering
    \includegraphics[width=0.9\textwidth]{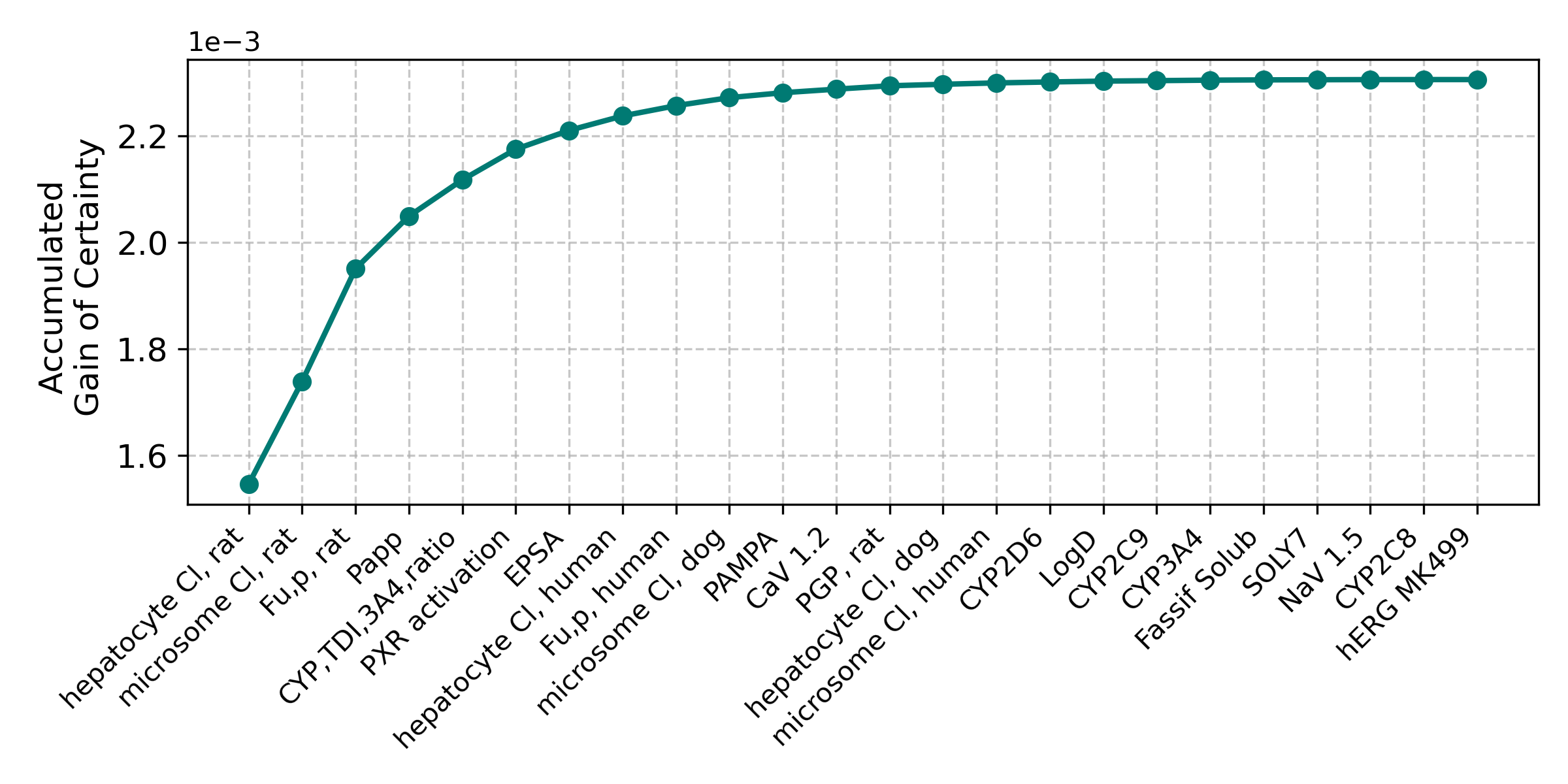}
    \caption{ Gain of certainty accumulated along the optimal (greedy) sequence of in vitro assays.  }
    \label{fig_gain_of_certainty}
\end{figure}

We illustrate this greedy scheme with the ADMET-750k dataset. We let ``MRT, rat'' be the assay of primary interest. We assume all in vivo experimental data (``halflife, rat'', ``Cl, rat'', ``Vd,rat'', ``halflife, dog'', ``MRT, dog'', ``Cl, dog'') is not available, and we allow all in vitro assays to be measured. Within the greedy scheme, we determine the optimal sequence of in vitro assays to be measured. The results, along with the accumulated GOC, are given in Fig.~\ref{fig_gain_of_certainty}. The accumulated GOC is the cumulative sum of the GOC of each new measurement along the sequence.

We find the top three assays in the optimal sequence are ``hypatocyte Cl, rat'', ``microsome Cl, rat'' and ``Fu,p, rat''. They contribute to more than $80\%$ of the final accumulated GOC. The types of the top three assays also align seamlessly with the empirical expectation that the in vitro properties directly associated with rat should efficiently improve the data completion of ``MRT, rat''. 
Compared to the top three assays, other in vitro assays bring only marginal GOC. The accumulated GOC saturates around the ``PAMPA'' assay.
Therefore, in practice, the termination of the experimental sequence can be set at any point between ``Fu,p, rat'' and ``PAMPA'', depending on budget and the cost of individual experiments. 

\section{Limitations}
A limitation of the current QComp approach is assuming all compounds in the database share the same covariance matrix. A compound-dependent covariance matrix may be introduced for a more fine-grained description of the probability density function in Eq.~\ref{dist}. 
The limitation of the performed benchmark is the lack of testing concurrent training of the base QSAR and QComp. Although the flexibility of using any existing QSAR model is a great advantage of QComp compared to integrated approaches such as 
Alchemite~\cite{whitehead2019imputation} and pQSAR~\cite{Martin2017}, it will be interesting to see if concurrent training can further improve the performance of QComp. In the future, a head-to-head comparison between Alchemite, pQSAR, and Chemprop-based QComp may be performed.


Moreover, the greedy scheme proposed for rational decision-making is limited by disregarding the fine-grained economic and ethical cost of each experiment. To achieve this, an objective function in terms of both GOC and the cost of each experiment should be designed. 

\section{Conclusions}\label{conclusions}

We have developed the QComp approach for reliable data completion. Having learned the intrinsic correlation between chemical activities, QComp is especially useful for instantaneously exploiting newly acquired sparse data for its own completion. At the same time, traditional QSAR approaches, including the multi-task ones~\cite{Heid2024}, can not absorb the knowledge from newly acquired data without retraining. 
We benchmarked QComp for ADMET data completion. QComp systematically improves upon structure-based QSAR models, such as Chemprop and random forest, and outperforms standard , iterative data-completion methods, including MICE, Missforest, and Macau, when they are all provided with the same side information. Notably, for assays where data completion approaches do not show an advantage over plain QSAR prediction, QComp yields similar $r^2$ scores as the base QSAR. Other data completion methods, however, may suffer from catastrophic failure. 

Then, we apply QComp to three major scenarios of drug discovery with favorable outcomes. 
First, QComp efficiently translates the knowledge from animal experiments to the prediction of human assays, improving $r^2$ scores obtained by bare QSAR prediction ($\approx 0.5$) to more than 0.7, a statistically significant figure for realistic applications. 
Second, QComp systematically improves upon conventional QSAR models for peptide drug discovery. In the future, QComp can be applied to material discovery where conventional QSAR models are also available~\cite{muratov2020qsar}. Third, QComp provides a concise and effective scheme for optimizing decision-making in preclinical drug discovery research, where acquiring in vivo assays is considerably more convenient than in vitro assays. 

These results demonstrate that QComp is accurate, robust, interpretable, and versatile. These advantages allow QComp to be integrated into most existing QSAR workflows of preclinical studies at a low cost. And we foresee more systematic, incremental applications of QComp.

\begin{ack}
We thank Ti-chiun Chang, Liying Zhang, and Alan Cheng for their insightful suggestions in preparing this manuscript. We are grateful to Merck \& Co. for supporting this work.
\end{ack}

\bibliography{references.bib}
\bibliographystyle{unsrt}



\newpage

\appendix


\renewcommand{\thetable}{S\arabic{table}}
\renewcommand{\thefigure}{S\arabic{figure}}
\setcounter{figure}{0}
\setcounter{table}{0}

\section{Model and training details}\label{sec_data_appendix}

\paragraph{Chemprop}

Chemprop consists of (1) a message passing network in which a graph structure of a molecule is transformed into a molecular latent representation and (2) a feed forward network which makes property predictions from the latent representation.  
A multi-task model is employed to predict all ADMET assays simultaneously as former studies have shown that multi-task models achieve better performance than single-task models when multiple tasks are correlated with each other \cite{feinberg2020improvement, Biswas2023PredictingLearning}. 
For the ADMET-750k dataset, the model is trained with an ensemble of 4 models, each initialized with a different random seed, and an epoch of 60. 10\% of the training set is randomly chosen as a validation set and used to determine the best epoch for the model during training. A hidden size of 600 and a depth of 4 are selected for the message-passing network. A hidden size of 1300 and a depth of 4 are selected for the feed-forward network. A normalized sum is used to aggregate the atomic embedding into a molecular embedding during the message-passing phase. For the public ADMET dataset, the models are trained using the same hyperparameters and ensembles with an epoch of 40.  

\paragraph{Random forest}
The random forest QSAR models for fup prediction were previously trained in-house on a larger, internal dataset. The random forest QSAR models for peptide prediction are trained on the peptide dataset. In both cases, a random forest model contains 500 trees and minimally 3 samples in a leaf node.  

\paragraph{QComp}
To train the QComp model for the ADMET-750k dataset, we let the total number of epochs be 4 and the batch size be 5000. We use the ADAM optimizer~\cite{kingma2014adam} for gradient descent in all our studies. Here, the initial learning rate is 0.003.  The learning rate decays by 0.5 every epoch. 
For the fup datasets, the number of epochs is 40 with a batch size of 5000.  The initial learning rate is 0.003. The learning rate decays by 0.5 every 15 epochs. For the peptide dataset, the number of epochs is 50 with a batch size of 1024. The initial learning rate is 0.01. The learning rate decays by 0.5 every epoch.
For the public ADMET dataset, the number of epoch is 10, with a batch size of 1000. We use an initial learning rate of 0.001. The learning rate decays by 0.5 every epoch. 

For all these datasets, the training of QComp can be accomplished by one multi-core Intel CPU within a few minutes or a few hours, depending on the size of the dataset. Specifically, training QComp for the public dataset takes less than one hour. For the test set of the public dataset, the data completion of one column takes less than 1 second.  

\paragraph{MICE} 
We use the IterativeImputer implemented in the fancyimpute~\cite{rubinsteyn2016fancyimpute} package for MICE~\cite{van2011mice} data completion. 
All parameters are default values (max\_iter=10, tol=0.001).

\paragraph{Missforest}

Missforest is an iterative imputation method similar to MICE. The difference is that the regression model in Missforest is random forest. 
In our study, we use the class ``IterativeImputer'' in the scikit-learn package for Missforest data completion. The regression model (estimator) is the random forest regressor (n\_estimators=4, max\_depth=10. max\_samples=0.5) in scikit-learn. The maximal iteration for iterative imputation is 25 with tol=0.1.

\paragraph{Macau}

Macau is a Bayesian probabilistic factorization method intended for sparse matrices analysis. Recently, Macau has been used for multi-task modeling in QSAR~\cite{simm2015macau}. In our study, the Python package Macau (v0.5.2) was used. The parameters are chosen as num\_latent=16, precision=5, burnin=400, and nsamples=1600. 

\section{Supplementary results}

\subsection{Proprietary ADMET-750k dataset with masked columns}\label{sec_masked_cols}

In this section, we address the realistic scenario in drug discovery that has not been reflected fully by the benchmark results in Sec.~\ref{sec_dataset1}. We consider the case that some highly correlated assays are either simultaneously present or simultaneously absent for arbitrary compounds. This corresponds to the situation in standardized laboratories where a certain group of assays are always measured at the same time from the same batch of samples. Such circumstances affect the imputation of missing values, especially for a target assay that all other assays highly correlated to it are missing at the same time. 

In the following, we intentionally create such circumstances with the ADMET-750k dataset. We will test if QComp can still yield reasonable performance. 
We adopt the scenario that the experimental data for a special group of assays, consisting of Mean Residence Time (MRT), half-life (halflife), clearance (Cl), and volume of distribution (Vd), are either present or absent simultaneously for the same species of animal. In our dataset, these assays are available for two species: dogs and rats, with the exception that Vd is not available for dogs. Here, we make a special protocol for incorporating this scenario in the data completion procedure: For imputing any assay in the special group for an animal species, we mask all columns of experimental data associated with the special group and the same species. For example, when we evaluate the performance of QComp on ``MRT, rat'', we mask the columns of ``MRT, rat'', ``halflife, rat'', ``Cl, rat'', and ``Vd, rat'' all together, while keeping the columns of ``MRT, dog'', ``halflife, dog'', ``Cl, dog'' unmodified. As for imputing assays outside the special group, we do not perform extra masking --- we still follow the protocol introduced in Sec.~\ref{sec_dataset1}.

\begin{figure}[h!]
    \centering
    \includegraphics[width=\textwidth]{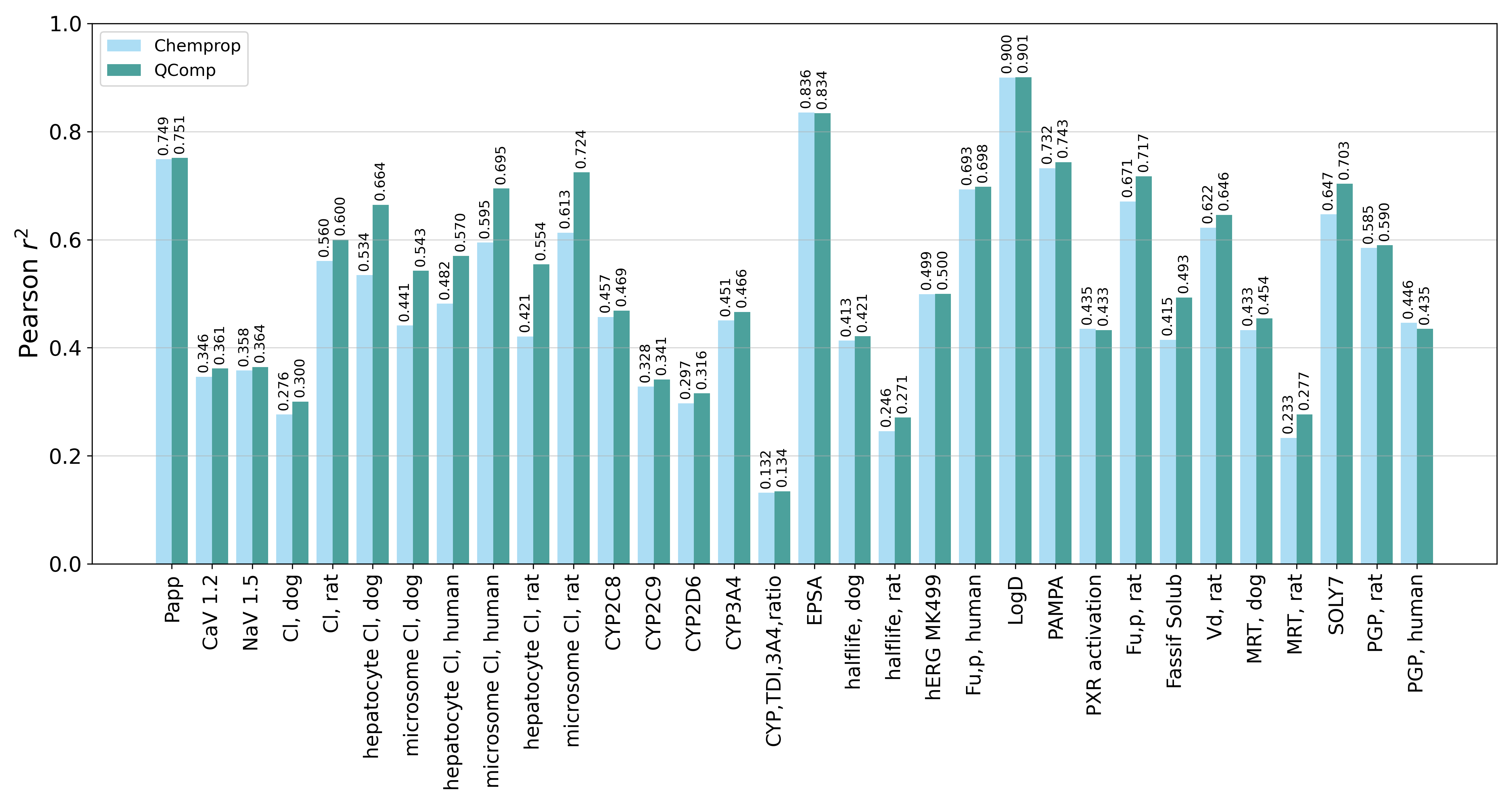}
    \caption{Performance of QComp and base QSAR model, Chemprop, on the masked ADMET-750k dataset (assay-based temporal splitting).}
    \label{fig_masked_cols}
\end{figure}

The performance of QComp against plain Chemprop prediction on the test set is shown in Fig.~\ref{fig_masked_cols}. Note that the results only differ from the QComp results in Table.~\ref{tab_dataset1} on the assays in the special group. Here, the $r^2$ score averaged over the special group (for both dogs and rats) is 0.398 for Chemprop and 0.424 for QComp. Previously, with the general protocol used to generate the results in Table.~\ref{tab_dataset1},  the $r^2$ score averaged over the special group is 0.853 for QComp. From this comparison, we find that QComp still brings a systematic improvement over base QSAR with the special protocol. 

\begin{figure}[h!]
    \centering
    \includegraphics[width=\textwidth]{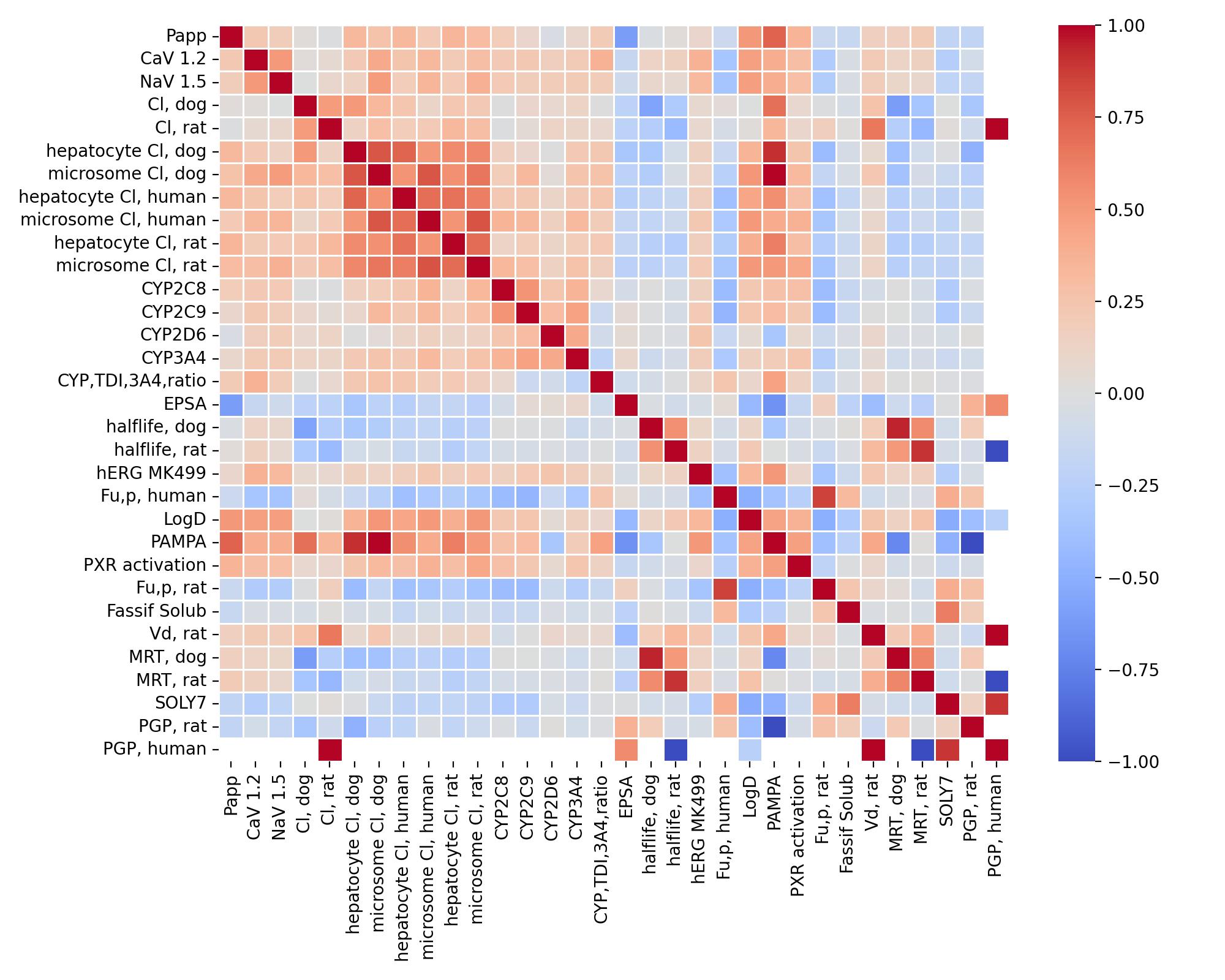}
    \caption{ADMET-750k dataset: Pearson correlation heatmap. Blank blocks indicate missing values (assays appearing mutually exclusively in the dataset). }
    \label{fig_dataset1_corr}
\end{figure}

Take the assay MRT as an example. For ``MRT, dog'', the improvement over base QSAR declines from 0.493 to 0.021. For  ``MRT, rat'', the improvement over base QSAR declines from 0.762 to 0.044. An explanation is elucidated by the assay-assay Pearson correlation heatmap plotted in Fig.~\ref{fig_dataset1_corr}, where MRT, dog shows an almost saturating (close to 1) correlation with halflife, dog, meanwhile, only weak correlation with other assays. Similarly, the MRT, rat is also highly correlated with the ``halflife, rat''. This saturating correlation can be understood from a rough exponential-decay model of Pharmacokinetics where a linear relation between MRT and half-life exists. With such a high correlation, under the circumstance that MRT is missing and halflife is present, one can impute MRT very accurately with QComp, as suggested by the previous results in Table.~\ref{tab_dataset1}. However, when MRT and halflife are both missing, the contribution of data completion becomes less significant, as is displayed here. Similar conclusions apply also to other assays in the special group.

\subsection{Public ADMET dataset}\label{sec_public}

\begin{figure}[h!]
    \centering
    \includegraphics[width=\textwidth]{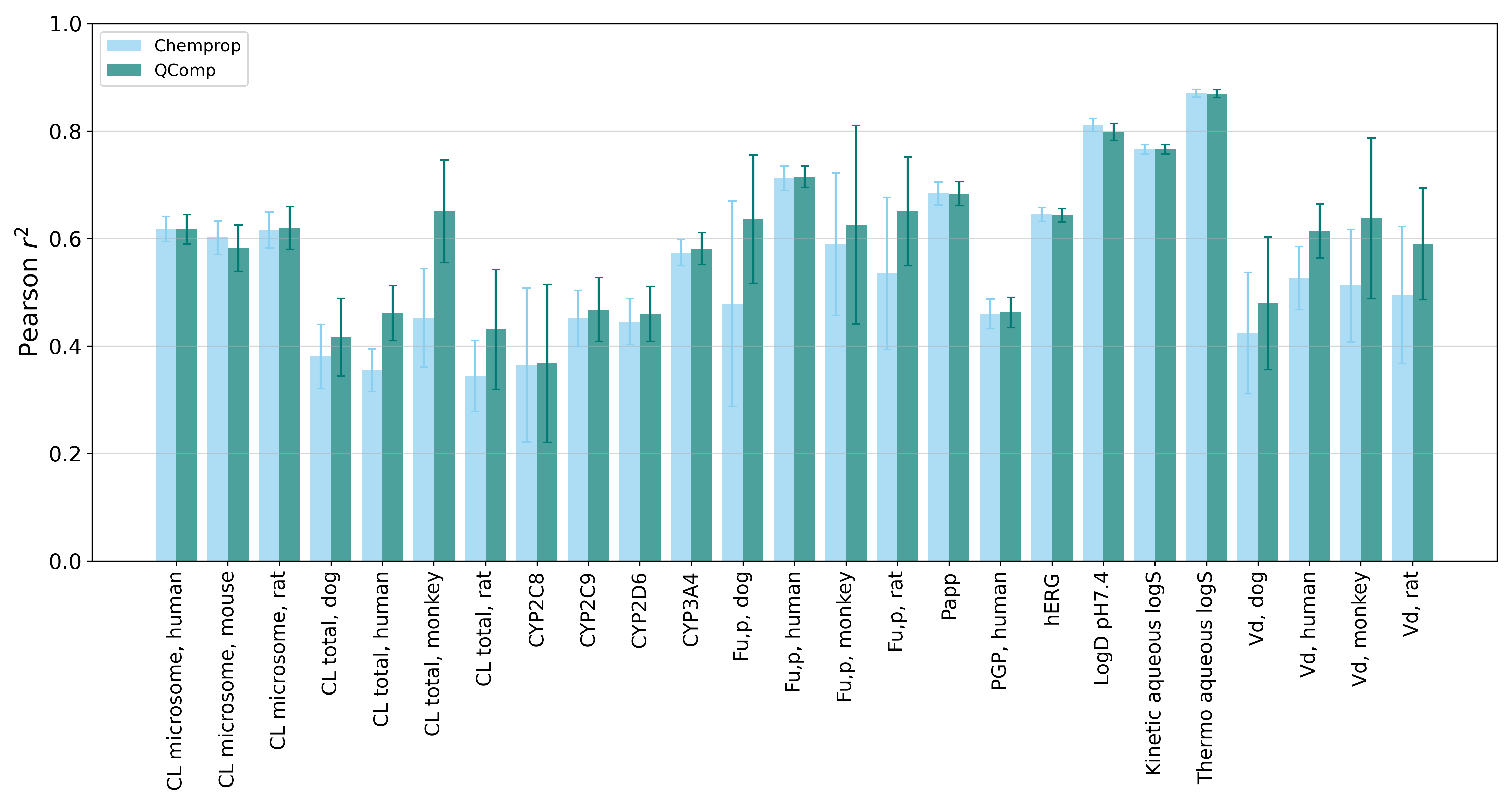}
    \caption{ Performance of QComp and base QSAR model (Chemprop) on the public dataset (random splitting). }
    \label{fig_public}
\end{figure}

We benchmark QComp on the public dataset, with the protocol introduced in Sec.~\ref{sec_dataset1}, for the reproducibility of this work. We also demonstrate whether the enhancement brought by QComp is robust over an ensemble of QSAR models trained on different splitting of the same dataset. 

We do a 5-fold random splitting (80\% training and 20\% test sets) of the public dataset. For each fold, we first train a Chemprop model as the base QSAR and then a QComp model with the same training set. Next, we evaluate the performance of QComp models on their respective test sets with the general protocol introduced previously. The results are given in Fig.~\ref{fig_public}, where the height of the bar and the associated error bar represent the 5-fold average and standard deviation of $r^2$ scores respectively. 
Here, the base QSAR models yield a mean $r^2$ score of 0.548, averaged over five folds and all assays. Meanwhile, the QComp models give a mean $r^2$ score of 0.593. Among all assays, CL total (clearance total), Fu,p (fraction unbound in plasma), and Vd (volume of distribution), associated with dog, human, monkey, and rat (12 assays in total), benefit considerably from QComp data completion with an average 0.092 gain in Pearson $r^2$ scores. For assays in this category, we find the lower end of the error bar (LEEB) associated with QComp is typically higher than or close to the bar associated with base QSAR, showing a robust advantage of QComp data completion. As for assays in the same category that do not gain significantly on $r^2$ scores, the QComp gives a LEEB higher than the LEEB from QSAR. The only exception is ``Fu,p, monkey'', where the LEEB of QComp is slightly lower than the LEEB of QSAR.
The assays not in this category, such as Cl microsome and Papp, do not receive considerable improvement from QComp. At the same time, no harm is done by QComp either --- the height of the bar and the size of the error bar have only negligible differences between QSAR and QComp. 

We conclude that QComp works on the public dataset also robustly and efficiently, without one case of catastrophic data completion displayed previously in Table.~\ref{tab_dataset1} by other methods. QComp is also robust against the deviation of base QSAR models trained on different splitting of the dataset.  Note that, the public dataset is compiled from multiple resources with potential inconsistency among data, which does not represent a typical use case of QComp in the industrial setting as reported in Sec.~\ref{sec_dataset1}.

\section{Composite Uncertainty}\label{sec_composite_uncertainty}

We let the uncertainty of $\boldsymbol{f}^{(i)}$ be denoted by $\boldsymbol{\sigma}^{(i)}=(\sigma_{1}^{(i)}, \sigma_{2}^{(i)},\cdots, \sigma_{p}^{(i)})$. In practice, both $\boldsymbol{f}^{(i)}$ and $\boldsymbol{\sigma}^{(i)}$ are calculated from an ensemble of deterministic QSAR models trained on the same dataset but initialized differently. $\boldsymbol{f}^{(i)}$ and $\boldsymbol{\sigma}^{(i)}$ are respectively the ensemble average and the standard deviation of QSAR predictions.
Assuming the components of $\boldsymbol{\sigma}^{(i)}$ are not correlated with each other, the ensemble covariance matrix associated with $\boldsymbol{f}^{(i)}$ is a diagonal matrix $\boldsymbol{\Sigma}_f^{(i)}$ with $(\sigma_j^{(i)})^2$ as $j$-th diagonal terms. Through propagation of uncertainty, the ensemble 
covariance matrix of $\boldsymbol{\mu}^{(i)}$ is $\boldsymbol{\Sigma}_{\boldsymbol{\mu}}^{(i)}=\mathbf{B}^{\top} \boldsymbol{\Sigma}_f^{(i)} \mathbf{B}$.
We can use $\boldsymbol{\Sigma}_{\boldsymbol{\mu}}^{(i)}$ to compute the ensemble deviation associated with $\tilde{\boldsymbol{\mu}}^{\mi}$. But before that, we need to define some extra notations. Let $(\mathbf{y}^{(i)})_j$ be an arbitrary missing assay and $(\mathbf{y}^{(i)})_k$ any known assay. $1\leq j,k \leq p$ are the indices of the assays in the whole collection of $p$ assays. In terms of the partition $\mathbf{y}^{(i)}=(\mathbf{y}^{\mi}, \mathbf{y}^{\oi})$, we use $j^{\mi}$ to denote the index of the assay-$j$ in the sub-vector $\mathbf{y}^{\mi}$, and $k^{\oi}$ the index of the assay-$k$ in $\mathbf{y}^{\oi}$. So there is an one-to-one mapping between $j$ and $j^{\mi}$, $k$ and  $k^{\oi}$. Additionally, we define $D^{(i)} = \boldsymbol{\Sigma}^{\moi} (\boldsymbol{\Sigma}^{\ooi})^{-1}$. So we can express the ensemble deviation associated to $\tilde{\boldsymbol{\mu}}^{\mi}$ in simple terms:
\begin{equation}
(\sigma_{\tilde{\boldsymbol{\mu}}^\text{M}}^{(i)})^2_{j^{\mi}} = (\boldsymbol{\Sigma}^{(i)}_\mu)_{jj} + \sum_{k^{\oi}=1}^{p^{(i)}_\text{O}} (D^{(i)}_{j^{\mi} k^{\oi} })^2 (\boldsymbol{\Sigma}^{(i)}_\mu)_{kk}.
\end{equation}
To incorporate the Gaussian statistical uncertainty assumed by QComp, we construct the composite uncertainty 
\begin{equation}
(\varsigma_{\tilde{\boldsymbol{\mu}}^\text{M}}^{(i)})^2_{j^{\mi}}  = (\sigma_{\tilde{\boldsymbol{\mu}}^\text{M}}^{(i)})^2_{j^{\mi}}  + (\widetilde{\boldsymbol{\Sigma}}^{\mmi})_{j^{\mi}j^{\mi}}.
\end{equation}
This final expression serves as a practical but rough estimation for the error of optimal data completion.

\newpage
\section{Dataset details}\label{sec_data_details}

\subsection{Proprietary ADMET-750k dataset}\label{sec_dataset1_details}

\newpage

For the assay-based split, each assay was considered independently and selected according to the date of the experiment.  

For the compound-based split, we split the dataset temporally according to the synthesis date of each compound. Because the training set of ``PGP, human'' by compound-based splitting is too small, with only 3 data points, this assay is not included in the experiment. 

Sec.~\ref{sec_dataset1} reports only the result on assay-based split.

\begin{table}[h!]
    \centering
    \caption{ADMET-750k dataset: Number of compounds in training and test sets for assay-based and compound-based temporal split}
    \scalebox{0.9}{
    \begin{tabular}{l lll lll}
    \toprule
        ~ & Assay-based & ~ & ~ & Compound-based & ~ & ~ \\  
        Assay & train size & test size & test size[\%] & train size & test size  & test size[\%] \\  
    \midrule
        
        Papp & 49542 & 5504 & 10.00 & 49272 & 5774 & 10.49 \\  
        CaV 1.2 & 138366 & 15373 & 10.00 & 142473 & 11266 & 7.33 \\  
        NaV 1.5 & 132700 & 14743 & 10.00 & 135994 & 11449 & 7.77 \\  
        Cl, dog & 19653 & 2183 & 10.00 & 19633 & 2203 & 10.09 \\  
        Cl, rat & 68310 & 7590 & 10.00 & 64395 & 11505 & 15.16 \\  
        hepatocyte Cl, dog & 7539 & 837 & 9.99 & 7232 & 1144 & 13.66 \\  
        microsome Cl, dog & 3972 & 441 & 9.99 & 3946 & 467 & 10.58 \\  
        hepatocyte Cl, human & 37974 & 4219 & 10.00 & 36476 & 5717 & 13.55 \\  
        microsome Cl, human & 43826 & 4869 & 10.00 & 44252 & 4443 & 9.12 \\  
        hepatocyte Cl, rat & 35258 & 3917 & 10.00 & 33531 & 5644 & 14.41 \\  
        microsome Cl, rat & 41265 & 4584 & 10.00 & 41609 & 4240 & 9.25 \\  
        CYP2C8 & 63305 & 7034 & 10.00 & 58548 & 11791 & 16.76 \\  
        CYP2C9 & 200590 & 22287 & 10.00 & 211790 & 11087 & 4.97 \\  
        CYP2D6 & 198458 & 22050 & 10.00 & 211776 & 8732 & 3.96 \\  
        CYP3A4 & 201351 & 22371 & 10.00 & 213576 & 10146 & 4.54 \\  
        CYP,TDI,3A4,ratio & 36293 & 4032 & 10.00 & 38477 & 1848 & 4.58 \\  
        EPSA & 36720 & 4080 & 10.00 & 18863 & 21937 & 53.77 \\  
        halflife, dog & 21498 & 2388 & 10.00 & 21541 & 2345 & 9.82 \\  
        halflife, rat & 74600 & 8288 & 10.00 & 70892 & 11996 & 14.47 \\  
        hERG MK499 & 327797 & 36422 & 10.00 & 349226 & 14993 & 4.12 \\  
        Fu,p, human & 20028 & 2225 & 10.00 & 19478 & 2775 & 12.47 \\  
        LogD & 413734 & 45967 & 10.00 & 457038 & 2663 & 0.58 \\  
        PAMPA & 3601 & 400 & 10.00 & 2907 & 1094 & 27.34 \\  
        PXR activation & 210816 & 23424 & 10.00 & 219501 & 14739 & 6.29 \\  
        Fu,p, rat & 49017 & 5446 & 10.00 & 43382 & 11081 & 20.35 \\  
        Fassif Solub & 284577 & 31620 & 10.00 & 247693 & 68504 & 21.66 \\  
        Vd, rat & 68329 & 7592 & 10.00 & 64431 & 11490 & 15.13 \\  
        MRT, dog & 17732 & 1970 & 10.00 & 17506 & 2196 & 11.15 \\  
        MRT, rat & 64805 & 7200 & 10.00 & 60538 & 11467 & 15.93 \\  
        SOLY7 & 424360 & 47150 & 10.00 & 412744 & 58766 & 12.46 \\  
        PGP, rat & 24868 & 2763 & 10.00 & 25214 & 2417 & 8.75 \\  
        PGP, human & 229 & 25 & 9.84 & 3 & 251 & 98.82 \\  
        \bottomrule
    \end{tabular}
    }
\end{table}

\newpage

\begin{table}[!ht]
    \centering
    \caption{ADMET-750k dataset: Performance of QComp and Chemprop for assay-based and compound-based temporal split}
    \begin{tabular}{l ll ll}
    \toprule
     ~ & Assay-based & ~  & Compound-based &~ \\  
        Assay & Chemprop  & QComp  & Chemprop  & QComp  \\ \midrule
        Papp & 0.749 & 0.751 & 0.721 & 0.725 \\  
        CaV 1.2 & 0.346 & 0.361 & 0.352 & 0.372 \\  
        NaV 1.5 & 0.358 & 0.364 & 0.347 & 0.368 \\  
        Cl, dog & 0.276 & 0.300 & 0.222 & 0.243 \\  
        Cl, rat & 0.560 & 0.600 & 0.387 & 0.430 \\  
        hepatocyte Cl, dog & 0.534 & 0.664 & 0.430 & 0.558 \\  
        microsome Cl, dog & 0.441 & 0.543 & 0.494 & 0.634 \\  
        hepatocyte Cl, human & 0.482 & 0.570 & 0.413 & 0.537 \\  
        microsome Cl, human & 0.595 & 0.695 & 0.472 & 0.619 \\  
        hepatocyte Cl, rat & 0.421 & 0.554 & 0.365 & 0.531 \\  
        microsome Cl, rat & 0.613 & 0.724 & 0.499 & 0.671 \\  
        CYP2C8 & 0.457 & 0.469 & 0.442 & 0.457 \\  
        CYP2C9 & 0.328 & 0.341 & 0.400 & 0.425 \\  
        CYP2D6 & 0.297 & 0.316 & 0.224 & 0.249 \\  
        CYP3A4 & 0.451 & 0.466 & 0.405 & 0.438 \\  
        CYP,TDI,3A4,ratio & 0.132 & 0.134 & 0.140 & 0.153 \\  
        EPSA & 0.836 & 0.834 & 0.816 & 0.812 \\  
        halflife, dog & 0.413 & 0.421 & 0.334 & 0.350 \\  
        halflife, rat & 0.246 & 0.271 & 0.224 & 0.237 \\  
        hERG MK499 & 0.499 & 0.500 & 0.470 & 0.474 \\  
        Fu,p, human & 0.693 & 0.698 & 0.596 & 0.649 \\  
        LogD & 0.900 & 0.901 & 0.837 & 0.847 \\  
        PAMPA & 0.732 & 0.743 & 0.494 & 0.482 \\  
        PXR activation & 0.435 & 0.433 & 0.384 & 0.387 \\  
        Fu,p, rat & 0.671 & 0.717 & 0.637 & 0.687 \\  
        Fassif Solub & 0.415 & 0.493 & 0.384 & 0.464 \\  
        Vd, rat & 0.622 & 0.646 & 0.582 & 0.613 \\  
        MRT, dog & 0.433 & 0.454 & 0.366 & 0.398 \\  
        MRT, rat & 0.233 & 0.277 & 0.165 & 0.181 \\  
        SOLY7 & 0.647 & 0.703 & 0.585 & 0.672 \\  
        PGP, rat & 0.585 & 0.590 & 0.494 & 0.502 \\  
        PGP, human & 0.446 & 0.435 & NaN & NaN \\  
        \bottomrule
    \end{tabular}
\end{table}

\newpage


\newpage
\subsection{The three-assay fup dataset}\label{sec_dataset2_details}

\begin{figure}[h!]
    \centering
    \includegraphics[width=0.4\textwidth]{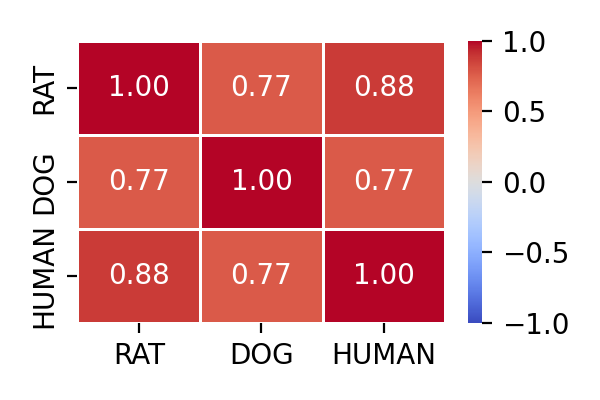}
    \caption{The fup dataset: Pearson correlation heatmap. }
    \label{fig_dataset2_corr}
\end{figure}

\begin{table}[!ht]
    \centering
    \caption{The fup dataset: Dataset size}
    \begin{tabular}{l lll}
    \toprule
        Assay & Data Count   \\  
        \midrule
        
        Rat& 48760 \\  
        Dog & 11711 \\  
        Human & 16883 \\  
        \bottomrule
    \end{tabular}
\end{table}

\newpage
\subsection{The peptide dataset}\label{sec_peptice_dataset}

\begin{figure}[h!]
    \centering
    \includegraphics[width=\textwidth]{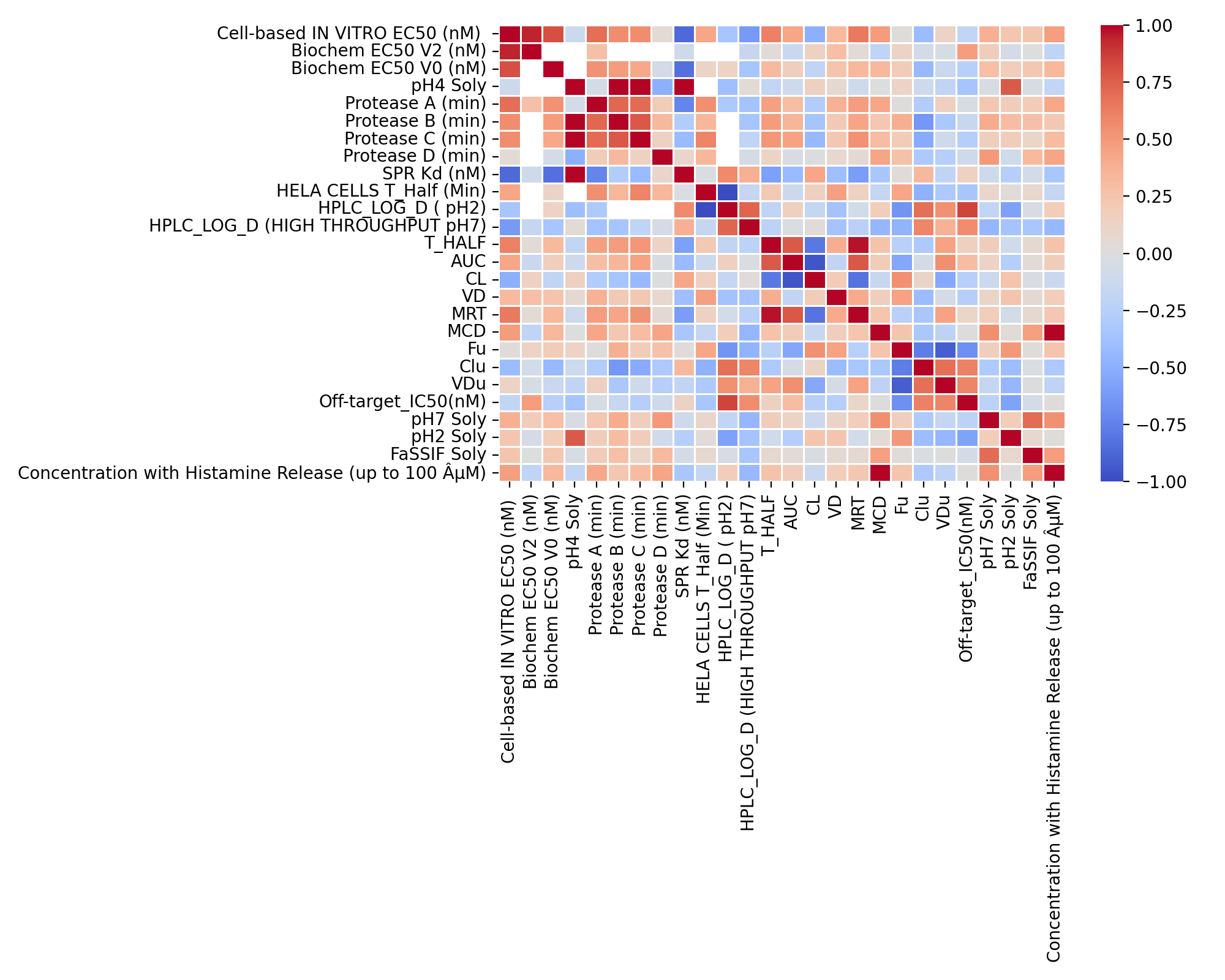}
    \caption{The peptide dataset: Pearson correlation heatmap. Blank blocks indicate missing values (assays appearing mutually exclusively in the dataset).}
    \label{fig_peptide_corr}
\end{figure}

\newpage
\subsection{The public dataset}\label{sec_public_details}

The public dataset (Sec.~\ref{sec_public}) used in this work is compiled from various public sources including Ref.~\citenum{Wenzel2019PredictiveSets} (ChEMBL, CC BY-SA 3.0 DEED), Ref.~\citenum{Iwata2022PredictingData} (CC-BY-NC-ND 4.0), PubChem~\citenum{Kim2023PubChemUpdate}, Ref.~\citenum{Watanabe2018PredictingRanges} (from PharmaPendium and ChEMBL), Ref.~\citenum{Falcon-Cano2022ReliableApproaches} (CC BY 4.0 DEED), Ref.~\citenum{Esposito2020CombiningSubstrates} (ChEMBL), Ref.~\citenum{Braga2015Pred-hERG:Toxicity}(ChEMBL), Ref.~\citenum{Aliagas2022ComparisonSets}(ChEMBL), Ref.~\citenum{Perryman2020PrunedSolubility}, Ref.~\citenum{Meng2022BoostingCuration}(CC BY 4.0 DEED), and Ref.~\citenum{Vermeire2022PredictingTemperatures}(CC BY 4.0 DEED).

Each assay data is converted to an appropriate unit as indicated in the Table~\ref{tab_public}. The SMILES identifiers from different data sources are validated and canonicalized using RDKit~\cite{Landrum2006RDKit:Cheminformatics}. The mean values are used when multiple data points are found for the same compound. 

\begin{figure}[h!]
    \centering
    \includegraphics[width=\textwidth]{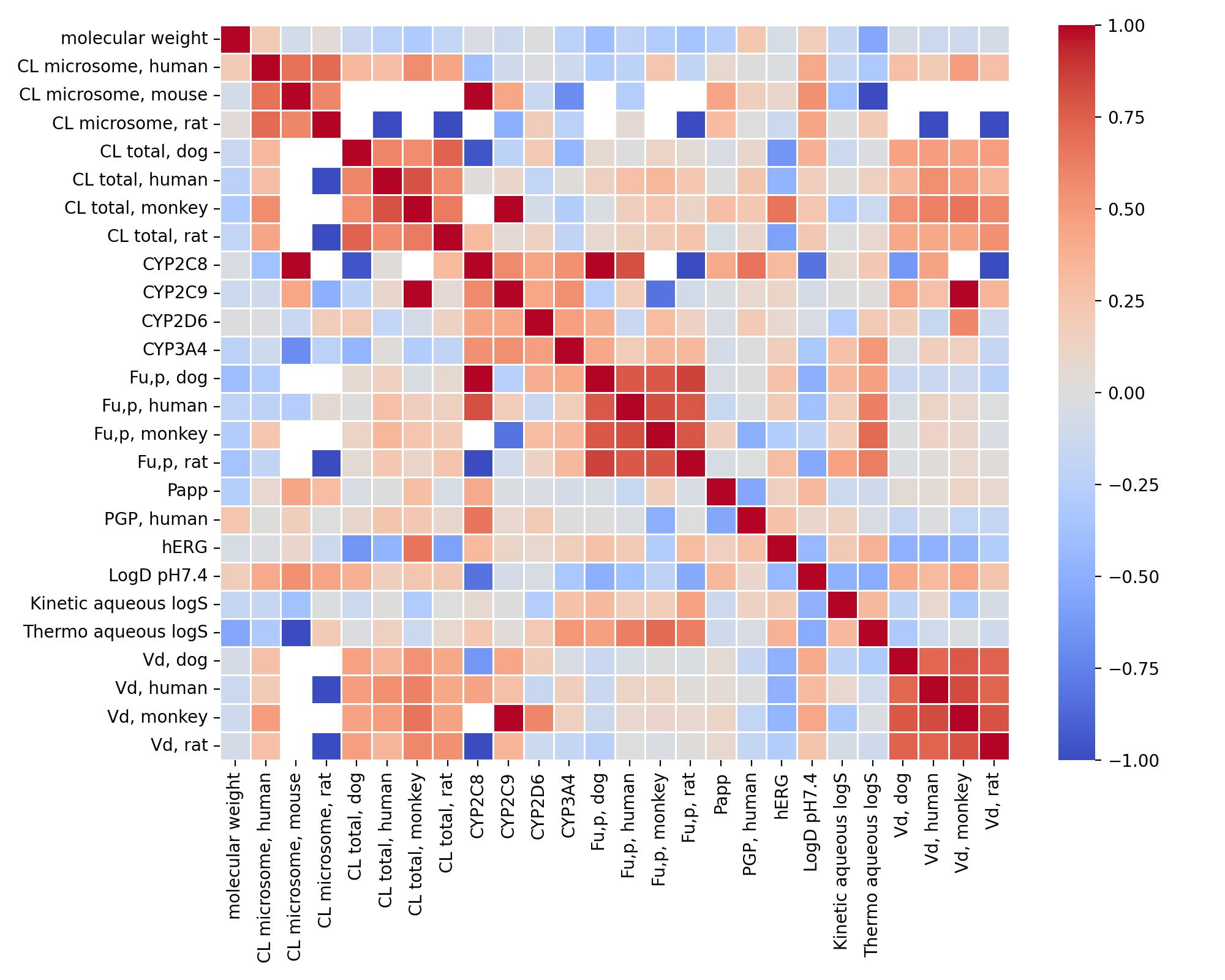}
    \caption{The public dataset: Pearson correlation heatmap. Blank blocks indicate missing values (assays appearing mutually exclusively in the dataset).}
    \label{fig_public_corr}
\end{figure}

\newpage
\begin{table}[!ht]
    \centering
    \caption{The public dataset: Dataset size, assay name map and units}
    \begin{tabular}{l l l l}
    \toprule
        Assay & Data Count & Short Name & Units \\  
        \midrule
        CL\_microsome\_human & 5218 & CL microsome, human & log10(mL/min/kg) \\  
        CL\_microsome\_mouse & 663 & CL microsome, mouse & log10(mL/min/kg) \\  
        CL\_microsome\_rat & 1798 & CL microsome, rat & log10(mL/min/kg) \\  
        CL\_total\_dog & 284 & CL total, dog & log10(mL/min/kg) \\  
        CL\_total\_human & 741 & CL total, human & log10(mL/min/kg) \\  
        CL\_total\_monkey & 129 & CL total, monkey & log10(mL/min/kg) \\  
        CL\_total\_rat & 387 & CL total, rat & log10(mL/min/kg) \\  
        CYP2C8\_inhibition & 328 & CYP2C8 & log10(nMolar IC50) \\  
        CYP2C9\_inhibition & 2374 & CYP2C9 & log10(nMolar IC50) \\  
        CYP2D6\_inhibition & 2539 & CYP2D6 & log10(nMolar IC50) \\  
        CYP3A4\_inhibition & 4403 & CYP3A4 & log10(nMolar IC50) \\  
        Dog\_fraction\_unbound\_plasma & 179 & Fu,p, dog & log10(fraction unbound) \\  
        Human\_fraction\_unbound\_plasma & 2717 & Fu,p, human & log10(fraction unbound) \\  
        Monkey\_fraction\_unbound\_plasma & 88 & Fu,p, monkey & log10(fraction unbound) \\  
        Rat\_fraction\_unbound\_plasma & 237 & Fu,p, rat & log10(fraction unbound) \\  
        Papp\_Caco2 & 6457 & Papp & log10($10^{-6}$ cm/s) \\  
        Pgp\_human & 2073 & PGP, human & log10(efflux ratio) \\  
        hERG\_binding & 5108 & hERG & log10(nMolar IC50) \\  
        LogD\_pH\_7.4 & 4190 & LogD pH7.4 & log10(M/M) \\  
        kinetic\_logSaq & 74895 & Kinetic aqueous logS & log10(M) \\  
        thermo\_logSaq & 11804 & Thermo aqueous logS & log10(M) \\  
        VDss\_dog & 274 & Vd, dog & log10(L/kg) \\  
        VDss\_human & 751 & Vd, human & log10(L/kg) \\  
        VDss\_monkey & 125 & Vd, monkey & log10(L/kg) \\  
        VDss\_rat & 351 & Vd, rat & log10(L/kg) \\  
        total\_compounds & 114112 & - & - \\  
        \bottomrule
    \end{tabular}
    \label{tab_public}
\end{table}



\end{document}